\documentclass[journal]{IEEEtran}
\pdfoutput=1
\usepackage[pdftex]{graphicx}
\usepackage{amsmath}
\usepackage{amssymb}

\usepackage{enumerate}

\usepackage{booktabs}
\usepackage{tabulary}
\usepackage{fixltx2e}
\usepackage[colorlinks=true,citecolor=black,urlcolor=blue,linkcolor=black]{hyperref}

\newcommand{\Section}[1]{\vspace{-4pt}\section{#1}\vspace{-4pt}}
\newcommand{\Subsection}[1]{\vspace{-3pt}\subsection{#1}\vspace{-3pt}}

\begin{document}

\title{Toward Long Distance, Sub-diffraction Imaging Using Coherent Camera Arrays}

\author{Jason~Holloway$^{1}$,
        M.~Salman~Asif$^1$,
				Manoj~Kumar~Sharma$^2$,
				Nathan~Matsuda$^2$,
				Roarke~Horstmeyer$^3$,
				Oliver~Cossairt$^2$,
				and~Ashok~Veeraraghavan$^1$% <-this % stops a space
\thanks{${}^1$J.~Holloway, M.S.~Asif, and A.~Veeraraghavan are with the Dept. of Electrical and Computer Engineering, Rice University, Houston, TX, 77005 USA e-mail: jh25@rice.edu}% <-this % stops a space
\thanks{${}^2$M.K.~Sharma, N.~Matsuda, and O.~Cossairt are with the Dept. of Electrical Engineering and Computer Science, Northwestern University, Evanston, IL, 60208.}% <-this % stops a space
\thanks{${}^3$R.~Horstmeyer is with the Dept. of Electrical Engineering, California Institute of Technology, Pasadena, CA, 91125.}
}

\IEEEpeerreviewmaketitle

\ifCLASSOPTIONcaptionsoff
  \newpage
\fi

\maketitle
%\thispagestyle{empty}

%%%%%%% ABSTRACT
\begin{abstract}
In this work, we propose using camera arrays coupled with coherent illumination as an effective method of improving spatial resolution in long distance images by a factor of ten and beyond.
Recent advances in ptychography have demonstrated that one can image beyond the diffraction limit of the objective lens in a microscope.
We demonstrate a similar imaging system to image beyond the diffraction limit in long range imaging.
We emulate a camera array with a single camera attached to  an X-Y translation stage. 
We show that an appropriate phase retrieval based reconstruction algorithm can be used to effectively recover the lost high resolution details from the multiple low resolution acquired images.
We analyze the effects of noise, required degree of image overlap, and the effect of increasing synthetic aperture size on the reconstructed image quality. 
We show that coherent camera arrays have the potential to greatly improve imaging performance.
Our simulations show resolution gains of $10\times$ and more are achievable.
Furthermore, experimental results from our proof-of-concept systems show resolution gains of $4\times - 7\times$ for real scenes.
Finally, we introduce and analyze in simulation a new strategy to capture macroscopic Fourier Ptychography images in a single snapshot, albeit using a camera array.
\end{abstract}

%%%%%%%% BODY TEXT
\Section{Introduction}
\label{sec:Intro}
Imaging from large stand-off distances typically results in low spatial resolution.
While imaging a distant target, diffraction blur, caused by the limited angular extent of the input aperture, is the primary cause of resolution loss.
As a consequence, it is desirable to use a large lens. 
However, telephoto lenses of comparable $f/\#$s are typically an order of magnitude more expensive and heavier than their portrait counterparts.
For example, consider imaging a subject at a distance of $z=1$ km with a lens aperture of $a = 12.5$ mm. 
Then the diffraction spot size is $\lambda z/a \approx 50$ mm diameter diffraction blur size at the object, completely blurring out important features, such as faces. 
If a large lens (same focal length but with a $125$ mm-wide aperture) is used instead, the diffraction blur is reduced to a $5$ mm diameter, which is small enough to be enable face recognition from $1$ km away.
Unfortunately, such telescopic lenses are expensive and bulky, and are thus rarely used.

\begin{figure}[t!]%
\centering
\includegraphics[width=\linewidth]{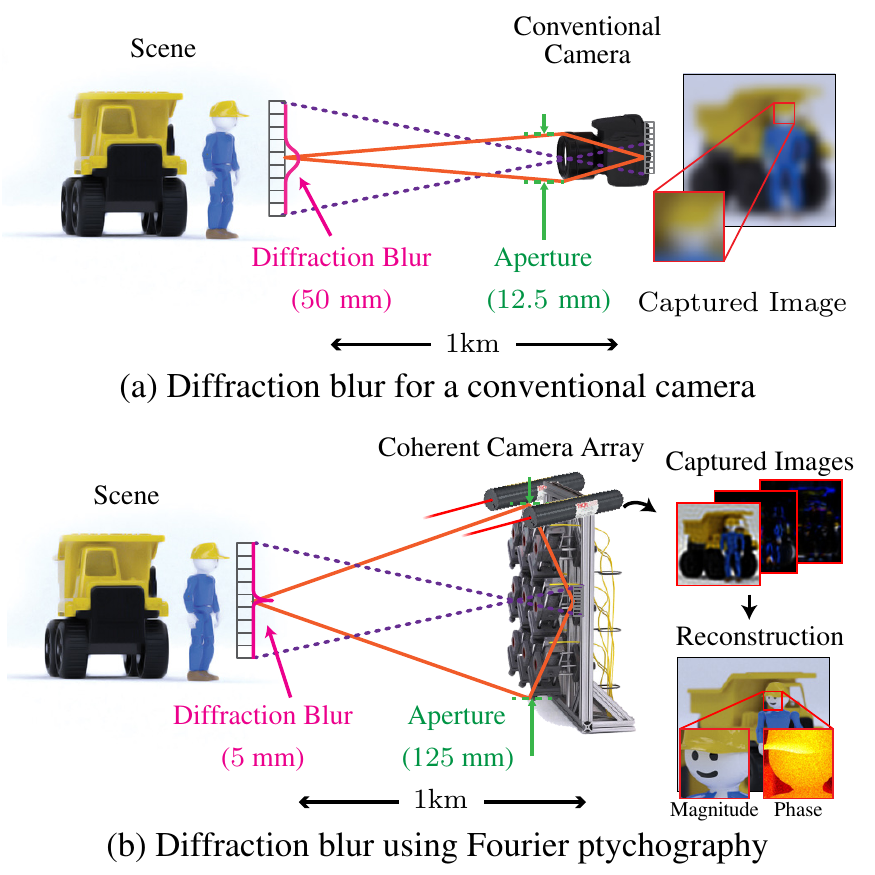}%
\caption{\textbf{Using active illumination to overcome the diffraction limit in long range imaging.} Diffraction blur is the primary cause of resolution loss in long distance imaging. Consider this illustration of imaging a human-sized object $1$ km away.
(a) A conventional camera using passive illumination with a fixed aperture size of $12.5$ mm induces a diffraction spot size of $50$ mm on objects $1$ km away, destroying relevant image features. (b) Using Fourier ptychography, an array of cameras using coherent illumination creates a synthetic aperture $10$ times larger than (a) resulting in a diffraction spot size of $5$ mm for scenes $1$ km away. Phase retrieval algorithms recover the high resolution image.}%
\label{fig:diffractionLimited}%
\end{figure}

\begin{figure*}%
\includegraphics[width=\linewidth]{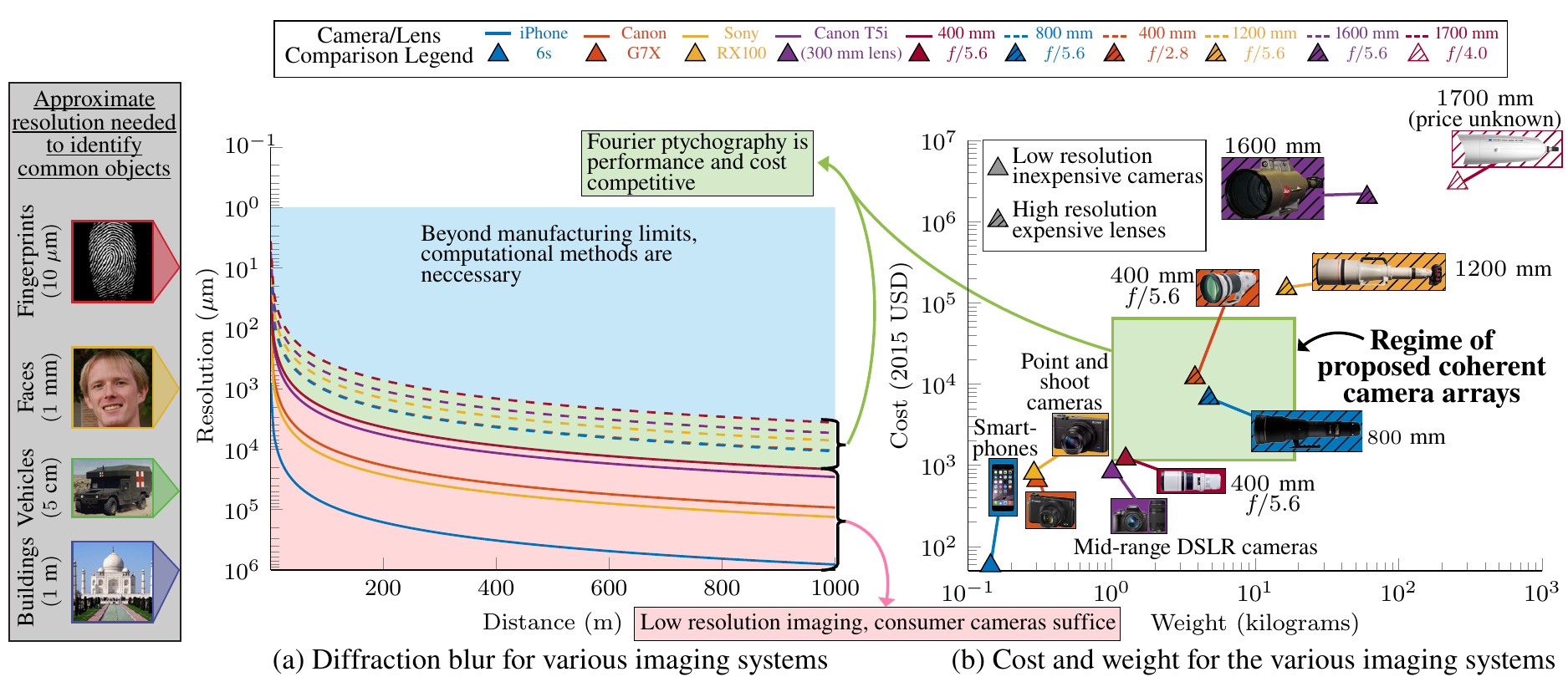}%
\caption{\textbf{Diffraction blur limits in long distance imaging.} (a) Diffraction blur spot size as a function of distance for common and exotic imaging systems. Consumer cameras (shaded in the pink region) are designed for close to medium range imaging but cannot resolve small features over great distances. Professional lenses (green region) offer improved resolution but are bulky and expensive. Identifying faces $1$ kilometer away is possible only with the most advanced super telephoto lenses, which are extremely rare and cost upwards of $\$2$ million US dollars \cite{walkenhorst2012world}. Obtaining even finer resolution (blue region) is beyond the capabilities of modern manufacturing; the only hope for obtaining such resolution is to use computational techniques such as Fourier ptychography. (b) Affordable camera systems (solid marks) are lightweight but cannot resolve fine details over great distances as well as professional lenses (striped marks) which are heavy and expensive. We propose using a camera array with affordable lenses and active lighting to achieve and surpass the capabilities of professional lenses. Note: These plots only account for diffraction blur and do not consider other factors limiting resolution (see \tablename~\ref{tab:motivTab} for more details).}%
\label{fig:motivFig}%
\vspace{-0.1in}
\end{figure*}

New computational imaging techniques are being developed to improve resolution by capturing and processing a collection of images. 
Here, we focus on one such multi-image fusion techniques designed specifically for the problem of recovering image details well below the diffraction limit.
It is convenient to perform multi-image fusion with a camera array, which can simultaneously acquire multiple images from different perspectives.
While camera arrays such as PiCam \cite{venkataraman2013picam} and Light\footnote{http://light.co} are becoming increasingly popular and have demonstrated resolution improvement, their applications are typically limited to improving resolution limits imposed by pixel sub-sampling.
They do not naturally extend to overcome the limits of diffraction blur.

Recent advances in ptychography have demonstrated that one can image beyond the diffraction-limit of the objective lens in a microscope \cite{zheng2013wide}.
Ptychography typically captures multiple images using a programmable coherent illumination source, and combines these images using an appropriate phase retrieval algorithm. 
It is also possible to implement ptychography by keeping the illumination source fixed, and instead spatially shifting the camera aperture \cite{dong2014aperture}.
Under this interpretation, ptychography closely resembles the technique advocated in this paper to improve resolution via a ``coherent camera array".
The goal of such a coherent array would be to recover both the amplitude and the phase of an optical field incident upon multiple spatially offset cameras, across the entire angular extent of the array. 
This matches the goal of creating a synthetic aperture via holography \cite{hillman2009high,martinez2008synthetic,massig2002digital}, but without directly measuring the phase of the incident field. 

In this paper, we explore the utility of ptychography-based methods for improving the imaging resolution in long-range imaging scenarios \cite{dong2014aperture}.
We analyze reconstruction performance under various scenarios to study the effects of noise, required degree of image overlap, and other important parameters and show that under practical imaging scenarios, coherent camera arrays have the potential to greatly improve imaging performance.
Further, we build a prototype and show real results of several subjects, demonstrating up to a $7\times$ quantitative improvement in spatial resolution.
Finally, we demonstrate that multiplexed illumination can be used to build a coherent camera array capable of capturing scenes in a single snapshot.

%%%%%
\Section{Resolution Limitations in Long Range Imaging}
\label{sec:ResLimits}
\begin{table*}[t]%
\centering
\caption{Intrinsic and extrinsic factors  which limit resolution in long range imaging}
\label{tab:motivTab}
\begin{tabular}{@{}llll@{}} \toprule
\normalsize{Resolution Limit}	&	\normalsize{Primary Cause}	&	\normalsize{Intrinsic/Extrinsic}	&	\normalsize{Example solutions}\\ \midrule \addlinespace[0.5em]
\textbf{Diffraction (this paper)}	&	\textbf{Finite aperture acceptance angle}	&	\textbf{Intrinsic}	&	\textbf{Synthetic aperture via holography, Ptychography} \\ \addlinespace[0.5em]
Sampling &	Finite Pixel Size	&	Intrinsic	&	Multi-image super-resolution~\cite{borman1998spatial}, dithering~\cite{fruchter2002drizzle}, \\
				 &										&						&	camera arrays~\cite{wilburn2005high}\\ \addlinespace[0.5em]
Aberrations	&	Finite lens track length/	&	Intrinsic	&	Additional optical elements, precalibration and \\
						& design complexity					&						&	digital removal~\cite{schuler2011non,kang2007automatic}, adaptive optics~\cite{booth2007adaptive}\\ \addlinespace[0.5em]
Noise & Sensor readout, shot noise & Intrinsic/Extrinsic	&	Image averaging, denoising~\cite{dabov2007image} \\ \addlinespace[0.5em]
Turbulence	&	Air temperature, wind shear	&	Extrinsic	&	Adaptive optics~\cite{ellerbroek1994first}, lucky imaging~\cite{fried1978probability,joshi2010seeing,roggemann1996imaging} \\ \addlinespace[0.5em]
Motion 	&	Camera and/or scene motion & Extrinsic	& Flutter shutter~\cite{holloway2012flutter,raskar2006coded}, Invariant blur~\cite{levin2008motion}\\\bottomrule
\end{tabular}
\end{table*}

Image resolution at long ranges is limited due to several inter-related factors as seen in \tablename~\ref{tab:motivTab}.
We classify these factors as being intrinsic to the imaging system (e.g. diffraction) and extrinsic to the imaging system (e.g. turbulence).

\textbf{Pixel Size:} 
A number of prior works developed techniques to overcome pixel-limited resolutions by capturing a sequence of images with sub-pixel translations \cite{borman1998spatial}. 
Some translation between the camera and scene is required between each exposure.
This is achieved using either small sensor movements, lens defocus, or movements of the entire imaging system \cite{baker2003shape}.
A sequence of sub-pixel shifted images may be acquired in a single snapshot by using a camera array.
However, current sensor pixels are approaching the diffraction limit of visible light ($1.2~\mu$m pixels are commercially widespread), and thus pixel sampling limits are not as critical a limitation to many current and future imaging platforms as in the past.

\textbf{Diffraction:}
Perhaps the most challenging impediment to imaging high resolution at a large distance is optical diffraction caused by the finite aperture size of the primary imaging lens.
While a wider lens will lead to a sharper point-spread function for a fixed focal length, it will also create proportionally larger aberrations. 
Additional optical elements may be added to the lens to correct these aberrations, but these elements will increase the size, weight and complexity of the imaging system \cite{lohmann1989scaling}. 
As noted in the introduction, the main focus of this work is to address this diffraction-based resolution limit with a joint optical capture and post-processing strategy. We use a small, simple lens system to capture multiple intensity images from different positions.
In our initial experiments our prototype uses a single camera on a translation stage that captures images sequentially (in approximately $90$ minutes).
The next phase of our research will focus on design, assembly, and calibration of a snapshot-capable camera array system.
We fuse captured images together using a ptychographic phase retrieval algorithm.
The result is a high resolution complex reconstruction that appears to have passed through a lens whose aperture size is equal to the synthetic aperture created by the array of translated camera positions.

\textbf{Additional causes of resolution loss:} We do not explicitly address the four remaining causes of resolution loss in this work.
We now mention their impact both for completeness and to highlight additional alternative benefits of the ptychographic technique.
First, intrinsic lens aberrations limit the resolution of every camera.
Second, turbulence is a large source of image degradation, especially when long exposure times are required.
Finally, noise and possible motion during the camera integration time leads to additional resolution loss. 

While not examined here, ptychography captures redundant image data that may be used to simultaneously estimate and remove microscope \cite{ou2014embedded} or camera \cite{horstmeyer2014overlapped} aberrations.
Redundant data is also helpful to address noise limits.
Furthermore, while yet to be demonstrated (to the best of our knowledge), is not hard to imagine that similar computational techniques may help estimate and remove external aberrations caused by turbulence, especially as our procedure simultaneously recovers the incident field's phase.
Ptychographic techniques thus appear ideally suited for overcoming the long-distance image resolution challenges summarized in \tablename~\ref{tab:motivTab}. 
In this paper, we address a limited problem statement; namely, can a ptychographic framework be extended to achieve sub-diffraction resolution at large standoff distances?

%%%%%
\Section{Related Work}
\label{sec:Related}
Computationally improving camera resolution is a longstanding goal in imaging sciences. Here, we summarize the prior work most relevant to our coherent camera array setup. 

\textbf{Multi-image super-resolution:}
Multi-image super-resolution has received much attention over the past $20$ years (see \cite{borman1998spatial} for a review of early algorithms) and continues to be an important area of study in modern camera arrays \cite{carles2014super,venkataraman2013picam}.
However, the overall improvement in resolution is has been modest; practical implementations achieve no more than a $2\times$ improvement \cite{baker2002limits,lin2004fundamental}.
Exemplar based efforts to improve the super-resolution factor beyond $2$ use large dictionaries to hallucinate information \cite{freeman2002example,glasner2009super} and offer no guarantees on fidelity.
Even these techniques are typically limited to producing a $4\times$ improvement in resolution.
In this paper, we present a technique that can, in theory, improve resolution all the way to the diffraction limit of light (e.g. below a micron for visible light).
Our experimental results clearly indicate significant resolution enhancements over and above those achieved by traditional multi-image techniques.

\textbf{Camera Arrays:}
Camera arrays comprised of numerous ($>\hspace{-1ex}10$) cameras, have been proposed as an efficient method of achieving a depth-of-field that is much narrower than that of a single camera in the array. 
Large arrays, such as the Stanford camera array \cite{wilburn2005high}, demonstrated the ability to computationally refocus.
Camera arrays are also well-suited to astronomical imaging. 
The CHARA array \cite{CHARA} is an collection of telescopes ($1$ meter diameter) on Mount Wilson in California which are jointly calibrated to resolve small astronomical features (angular resolution of $200$ microarcseconds), using a synthetic aperture with a diameter of $330$ meters.

Despite the advantages provided using a camera array, diffraction blur limits resolution in long distance imaging.
For scenes lit with incoherent illumination, such as sunlight, multi-image super-resolution techniques can further increase spatial resolution by a factor of two.
Even with this modest improvement in resolution, the remaining blur due to diffraction may still preclude recovering fine details.
In this paper we propose using active illumination to recover an image with spatial resolution many times greater than the limit imposed by optical diffraction.

\textbf{Fourier Ptychography}
Fourier ptychography (FP) has permitted numerous advances in microscopy, including
wide-field high-resolution imaging~\cite{zheng2013wide}, removing optical~\cite{horstmeyer2014overlapped} and pupil~\cite{ou2014embedded} aberrations, digital refocusing~\cite{dong2014aperture}, and most recently $3$D imaging of thick samples~\cite{li2015separation,tian20153d}.
FP is, in effect, a synthetic aperture technique~\cite{hillman2009high,martinez2008synthetic,massig2002digital,mico2006synthetic} that does not require phase measurements.
In FP microscopy, the imaging lens is often fixed while an array of light sources is used to shift different regions of the image spectrum across a fixed aperture of limited size.
Instead of advancing the light in a raster pattern, the illumination sources may be multiplexed to increase SNR and reduce acquisition time~\cite{dong2014spectral,tian2014multiplexed}.
Refinements in the sampling pattern can also reduce acquisition time~\cite{guo2015optimization}.
Initial work in macroscopic imaging~\cite{dong2014aperture} suggests the potential benefits of Fourier ptychography in improving spatial resolution of stand-off scenery.
In this paper we extend upon this work and characterize how the amount of overlap, synthetic aperture size, and image noise affects reconstruction quality and resolution improvement.
Furthermore, we describe early efforts towards realizing a physical system to beat diffraction in long range snapshot imaging systems.% (e.g., coherent camera arrays).

\textbf{Phase Retrieval Algorithms:}
Phase retrieval is a required component of ptychographic image reconstruction.
Image sensors record the magnitude of a complex field, either in the Fourier domain for traditional ptychography or the spatial domain in Fourier ptychography.
Numerous algorithms for retrieving the missing phase information have been proposed. One of the simplest techniques is based upon alternating projections~\cite{fienup1982phase}.
Many extensions of this simple starting point exist for both standard phase retrieval using a single image~\cite{elser2003phase,fienup1986phase}, as well as ptychographic phase retrieval using multiple images~\cite{horstmeyer2015convex,maiden2009improved}.

\textbf{Alternative methods to improve resolution:}
The primary technique to improve image resolution in long-distance imaging is to simply build a larger telescope.
On the scale of portable imaging systems, this typically means removing large glass optical elements in favor of catadioptric lens designs, where a mirror reflects light through a much smaller lens.
Other techniques for improving resolution include lucky imaging to mitigate atmospheric turbulence~\cite{fried1978probability,joshi2010seeing,zhang2011efficient} or motion tracking in astronomical imaging~\cite{dantowitz2000ground,mcclure1989image}.
These latter approaches tackle external factors limiting resolution, but do not address the effects of diffraction on spatial resolution.

%%%%%
\Section{Fourier Ptychography for Long Range Imaging}
\label{sec:fourierPtych}
Our goal in this paper is to capture high resolution images of objects that are a considerable distance away from the camera, e.g. resolve a face $1000$ meters away.
Passive camera systems use available (incoherent) illumination sources such as the sun and suffer from significant diffraction blur.
Existing super resolution methods may be able to effectively reduce the amount of blur by a factor of two, which may be insufficient to resolve enough detail to recover the object of interest.
In order to resolve even smaller features, we assume that we have a coherent or partially coherent illumination source; that is, we have active illumination.

Like other multi-image super-resolution methods \cite{baker2002limits,lin2004fundamental} we will capture a series of low resolution images which we will then use to recover a high resolution image.
We take multiple images using a coherent illumination source where each image is from a different (known) position in the $XY$ plane.
For simplicity we assume the camera positions coincide with a regular grid, though this is not necessary \cite{guo2015optimization}.
It should be noted that we would obtain the same result by leaving the camera stationary and moving the illumination source, a common practice in microscopy~\cite{dong2014spectral,horstmeyer2015convex,ou2014embedded,tian2014multiplexed,tian20153d}.

\Subsection{Image Formation Model}
We first assume a single fixed illumination source\footnote{In Section~\ref{sec:prelimWork} we consider multiple illumination sources}, which can either be co-located with or external to our imaging system.
The source should be quasi-monochromatic, with center wavelength $\lambda$.
For color imaging, multiple quasi-monochromatic sources (i.e., laser diodes) are effective~\cite{zheng2013wide}.
The source emits a field that is spatially coherent across the plane which contains our object of interest, $P(x,y)$.
We assume this plane occupies some or all of the imaging system field of view. 

The illumination field, $u(x,y)$, will interact with the object, and a portion of this field will reflect off the object towards our imaging system.
For our initial experiment, we assume the object is thin and may be described by the $2$D complex reflectivity function $o(x,y)$.
Extension to surface reflectance from $3$D objects is straightforward\footnote{Extension to complex materials such as densely scattering media is significantly more complicated.}.
Under the thin object approximation, the field emerging from the object is given by the product $\psi(x,y)=u(x,y)o(x,y)$.
This then propagates a large distance $z$ to the far field, where our imaging system occupies the plane $S(x',y')$. 

Under the Fraunhofer approximation, the field at $S(x',y')$ is connected to the field at the object plane by a Fourier transform:
\begin{equation}
\widehat{\psi}(x',y') = \frac{e^{jkz}e^{\frac{jk}{2z}(x'^2+y'^2)}}{j\lambda z} \mathcal{F}_{1/\lambda z} \left[\psi(x,y) \right]
\label{ft}
\end{equation}
where $k=2\pi/\lambda$ is the wavenumber and $\mathcal{F}_{1/\lambda z}$ denotes a two dimensional Fourier transform  scaled by $1/\lambda z$.
For the remainder of this manuscript, we will drop multiplicative phase factors and coordinate scaling from our simple model.
We note that the following analysis also applies under the Fresnel approximation (i.e., in the near field of the object), as supported by recent investigations in X-ray ptychography~\cite{edo2013sampling,thibault2008high}.

The far field pattern, which is effectively the Fourier transform of the object field, is intercepted by the aperture of our camera.
We describe our limited camera aperture using the function $A(x'-c_{x'},y'-c_{y'})$, which is centered at coordinate $(c_{x'},c_{y'})$ in the plane $S(x',y')$ and passes light with unity transmittance within a finite diameter $d$ and completely blocks light outside (i.e., it is a ``circ" function). 

The optical field immediately after the aperture is given by the product $\widehat{\psi}(x',y') A(x'-c_{x'},y'-c_{y'})$.
This bandlimited field then propagates to the image sensor plane.
Again neglecting constant multiplicative and scaling factors, we may also represent this final propagation using a Fourier transform.
Since the camera sensor only detects optical intensity, the image measured by the camera is 
\begin{equation}
I(x,y,c_{x'},c_{y'})\propto \left|\mathcal{F}\left[ \widehat{\psi}(x',y') A(x'-c_{x'},y'-c_{y'}) \right]\right|^{2}.
\label{image}
\end{equation}
In a single image, the bandpass nature of the aperture results in a reduced resolution image.
For an aperture of diameter $d$ and focal length $f$, diffraction limits the smallest resolvable feature within one image to be approximately $1.22\lambda f/d$. 

\begin{figure}[t!]%
\centering
\includegraphics[width=\columnwidth]{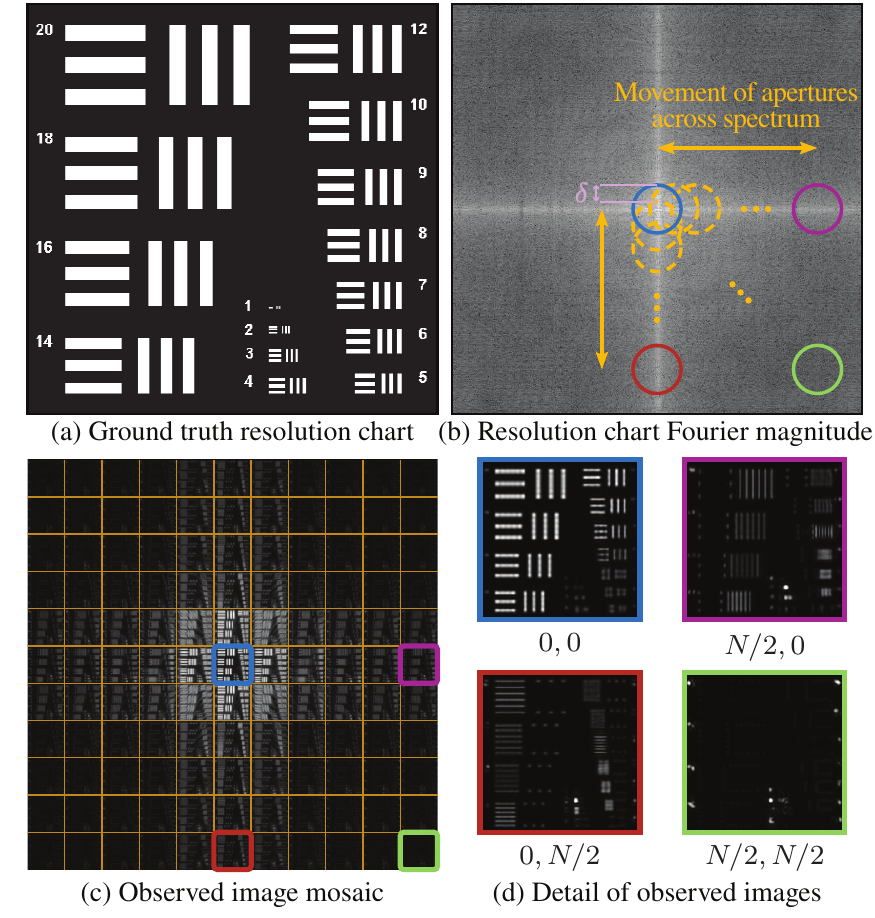}
\caption{\textbf{Sampling the Fourier domain at the aperture plane.} For a given high resolution target (a), the corresponding Fourier transform is formed at the aperture plane of the camera (b). The lens aperture acts as a bandpass filter, only allowing a subset of the Fourier information to pass to the sensor. A larger aperture may be synthesized by scanning the aperture over the Fourier domain and recording multiple images. (c) The full sampling mosaic acquired by scanning the aperture. The dynamic range has been compressed by scaling the image intensities on a log scale. (d) Larger detail images shown of four camera positions, including the center. Image intensities have been scaled linearly, note that only high frequency edge information is present in the three extreme aperture locations. Please view digitally to see details.}%
\label{fig:FPsampling}%
\end{figure}

\Subsection{Fourier Ptychography to Improve Resolution}
Ptychography presents one strategy to overcome the diffraction limit by capturing multiple images and synthetically increasing the effective aperture size.
The series of captured images are used to recover the high resolution complex field in the aperture plane and subsequently a high resolution image.

To achieve this, we re-center the camera at multiple locations, $\left(c_{x'}(i),c_{y'}(i)\right)$, and capture one image at the $i$th camera location, for $i=1,\ldots,N$.
This transforms \eqref{image} into a four-dimensional discrete data matrix.
The $N$ images can be captured in a number of ways, one can: physically translate the camera to $N$ positions, construct a camera array with $N$ cameras to simultaneously capture images, fix the camera position and use a translating light source, or use arbitrary combinations of any of these techniques.

If we select our aperture centers such that they are separated by the diameter $d$ across a rectilinear grid, then we have approximately measured values from the object spectrum across an aperture that is $\sqrt{N}$ times larger than what is obtained by a single image.
Thus, it appears that such a strategy, capturing $N$ images of a coherently illuminated object in the far field, may be combined together to improve the diffraction-limited image resolution to $1.22\lambda f/\sqrt{N}d$.

However, since our detector cannot measure phase, this sampling strategy is not effective as-is.
Instead, it is necessary to ensure the aperture centers overlap by a certain amount (i.e., adjacent image captures are separated by a distance $\delta<d$ along both $x'$ and $y'$).
This yields a certain degree of redundancy within the captured data, which a ptychographic post-processing algorithm may utilize to simultaneously determine the phase of the field at plane $S(x',y')$.
Typically, we select $\delta\sim0.25d$.
See \figurename~\ref{fig:FPsampling} for an overview of the sampling strategy and example images recorded with the aperture acting as a bandpass filter of the Fourier transform $\widehat{\psi}(x',y')$.
Next, we detail a suitable post-processing strategy that converts the data matrix $I\left(x,y,c_{x'}(i),c_{y'}(i)\right)$ into a high-resolution complex object reconstruction.

%%%%%
\Section{Algorithm for Image Recovery}
\label{sec:imRecovAlgo}
\begin{figure}[t!]%
\includegraphics[width=\linewidth]{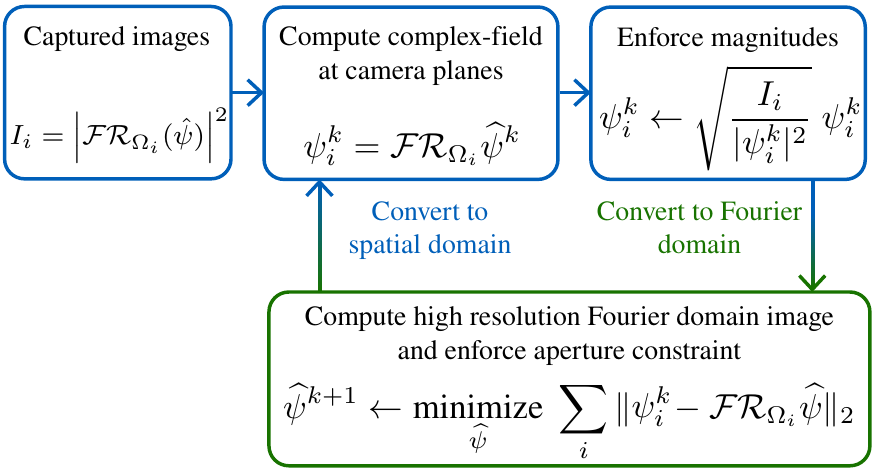}
\caption{\textbf{Block diagram of the image recovery algorithm.} Constraints on the image domain magnitude and Fourier domain support are enforced in an alternating manner until convergence or a maximum iteration limit is met.}%
\label{fig:AM_blocks}%
\end{figure}

Using our coherent camera array measurements in \eqref{image}, our goal is to recover the complex-valued, high-resolution field $\widehat \psi(x',y')$. This computational goal approximately matches the goal of ptychography. Let us denote the image measured with camera at location $(c_{x'_i},c_{y'_i})$ as
\begin{equation}
I_i = |\psi_i|^2\equiv|\mathcal{F}\mathcal{R}_{\Omega_i} \widehat \psi|^2,
\end{equation}
where $\psi_i=\mathcal{F}\mathcal{R}_{\Omega_i} \widehat \psi$ denotes the complex-valued, bandlimited field whose intensity is measured at the image sensor, $\widehat \psi$ denotes the complex-valued field at the Fourier plane (i.e. the aperture plane), and $\mathcal{R}_{\Omega_i} \widehat \psi$ denotes an aperture operator that sets all the entries of $\widehat \psi(x',y')$ outside the set $\Omega_i = \{(x,y): |x-c_{x'_i}|^2 + |y-c_{y'_i}|^2 \le d/2\}$ to zero. 

To recover the high-resolution $\widehat\psi$ from a sequence of $N$ low-resolution, intensity measurements, $\{I_i\}_{i=1}^N$, we use an alternating minimization-based phase retrieval problem that is depicted in \figurename~\ref{fig:AM_blocks}.
We seek to solve the following problem
\begin{equation}
\widehat\psi^* = \underset{\widehat\psi}{\operatorname{arg~min}} \sum_i{\left\|\psi_i - \mathcal{FR}_{\Omega_i} \widehat\psi\right\|_2}\;\; \textnormal{s.t.}\;\;|\psi_i|^2 = I_i,
\label{eq:problem1}
\end{equation}
by alternatively constraining the support of $\widehat\psi$ and the squared magnitudes of $\psi$.
We set the initial estimate of $\widehat\psi^0$ to be a scaled version of the Fourier transform of the mean of the low resolution images.
At every iteration ($k$), we perform the following three steps: 
\begin{enumerate}[\bf 1.]
	\item  Compute complex-valued images at the sensor plane using the existing estimate of the field at the Fourier plane, $\widehat \psi^k$:
	$$\psi_i^k = \mathcal{FR}_{\Omega_i} \widehat \psi^k \quad \text{for all } i.$$
	\item  Replace the magnitudes of $\psi_i^k$ with the magnitude of the corresponding observed images $I_i$: 
	$$ \psi_i^k \gets \sqrt{\frac{I_i}{|\psi_i^k|^2}}\; \psi_i^k \quad \text{for all } i.$$
	\item  Update the estimate of $\widehat \psi$ by solving the following regularized, least-squares problem: 
	\begin{equation}
	\widehat \psi^{k+1} \gets \underset{\widehat \psi}{\textnormal{minimize}}\; \sum_i \left\|\psi_i^k - \mathcal{FR}_{\Omega_i} \widehat \psi\right\|_2^2+\tau\|\widehat \psi\|_2^2,
	\end{equation}
	where $\tau>0$ is an appropriately chosen regularization parameter. This problem has a closed form solution, which can be efficiently computed using fast Fourier transforms. 
\end{enumerate}

%%%%%
\Section{Performance Analysis and Characterization}
\label{sec:simRecov}
Accurate retrieval of optical phase requires redundant measurements. If the final high-resolution image is comprised of $n$ pixels that contain both intensity and phase information, then it is clear that we must acquire at least $2n$ measurements of optical intensity from the image sensor. However, it is not clear if additional data should be acquired (for example to improve tolerance to noise), how much this additional data might improve the quality of our reconstructions a different resolution scales, or how many images on average might be required to achieve a certain resolution goal. We now explore these questions via simulation.

\Subsection{Fundamental Factors that Affect Recovery Performance}
There are two parameters that will influence the number of acquired images. First is a desired resolution limit of our coherent camera array (i.e., its minimum resolvable feature), which is equivalent to specifying the synthetic aperture size. Second is the desired degree of data redundancy, which is equivalent to specifying the amount of overlap between adjacent images when viewed in the Fourier domain. Computational resources are proportional to the number of acquired images, so the user will be eventually forced to trade off reconstruction quality for reconstruction time.

\Subsection{Experimental Design}
To explore these two parameters, we perform experimental simulations using a $512$ px $\times 512$ px resolution chart shown in \figurename~\ref{fig:FPsampling}(a).
This chart contains line pairs with varying widths from $20$ pixels down to $1$ pixel, corresponding to line pairs per pixel in the range $[0.025,0.5]$.
We assume each object is under coherent plane wave illumination from a distant source.
Diffraction blur is determined solely by the wavelength of the coherent source and the ratio between the focal length and aperture diameter (i.e. the $f/\#$). 
We assume that the illumination wavelength is $550$ nm, the focal length of the lens is $800$ mm, and the aperture diameter is $18$ mm.
The resolution chart itself is $64$ mm $\times 64$ mm and is located $50$ meters away from the camera.
The aperture of the imaging system is scanned across the Fourier plane to generate each bandpassed optical field at the image plane.
A pictorial representation of the sampling pattern and resulting captured images is shown in \figurename~\ref{fig:FPsampling}(b)-(d).

For simplicity, we treat the target resolution chart as an amplitude object, which may be considered to be printed on a thin glass substrate or etched out of a thin opaque material.
We use the iterative algorithm described in Section~\ref{sec:imRecovAlgo} to recover the complex field of the resolution chart.
The algorithm terminates whenever the relative difference between iterations falls below $10^{-5}$ or after $1000$ iterations.
To quantify the reconstruction we compute the contrast for the horizontal and vertical bars belonging to each group.
Contrast, $C$, is defined as 
\begin{equation}
C = \frac{\overline{w} - \overline{b}}{\overline{w} + \overline{b}},
\label{eq:contrast}
\end{equation}
where $\bar{w}$ and $\bar{b}$ denote the average intensity of the white and black bars respectively. 
To aid our discussion, we define the limit of resolvability to be when the contrast of a group drops below $20\%$ (MTF$20$).
For the simulation experiments, we will assume the image sensor has a pixel width of $2~\mu$m.
It is important to note that in the simulation experiments we will only consider the effects of diffraction blur as the factor which limits resolution in long range images.

\begin{figure}[t!]%
\centering
\includegraphics[width=\linewidth]{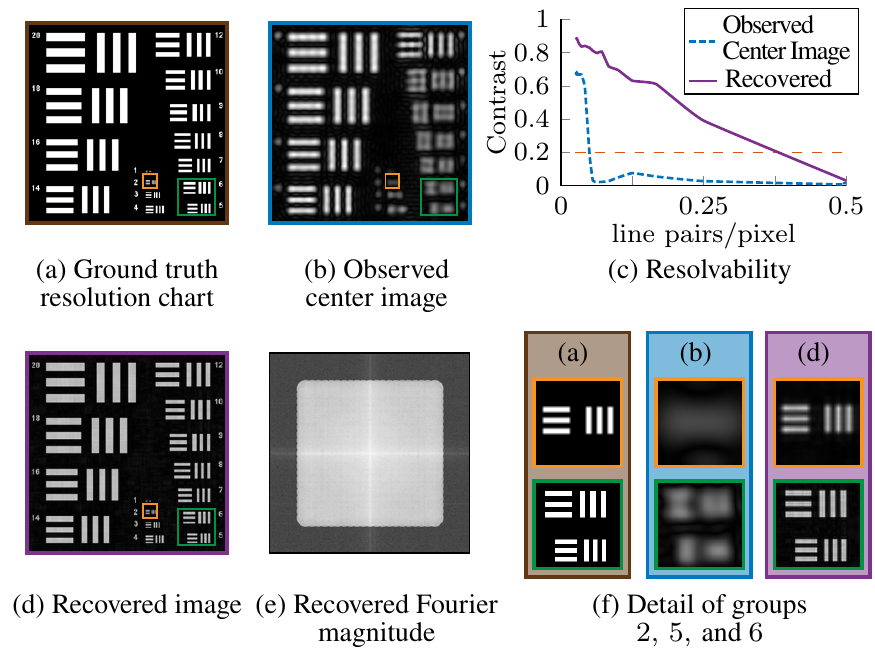}
\caption{\textbf{Recovering high frequency information using a Fourier ptychography.}
(a) We simulate imaging a $64 \times 64$ mm resolution target $50$ meters away using sensor with a pixel pitch of $2~\mu$m. The width of a bar in group $20$ is $2.5$ mm.
(b) The target is observed using a lens with a focal length of $800$ mm and an aperture of $18$ mm. The aperture is scanned over a $21\times21$ grid ($61\%$ overlap) creating a synthetic aperture of $160$ mm. The output of phase retrieval is a high resolution Fourier representation of the target.
The recovered image is shown in (d) and the recovered Fourier magnitude (log scale) is shown in (e).
The plot in (c) shows the contrast of the groups in the intensity images of the recovered resolution chart (purple solid line) and the observed center image (blue dashed line).
Whereas the central image can only resolve elements which have a width of $12$ pixels before contrast drops below $20\%$, using Fourier ptychography we are able to recover features which are only $2$ pixels wide.
Detail images of groups 2, 5, and 6, for the ground truth, observed, and recovered images are shown in (f).}%
\label{fig:simResChart}%
\end{figure}

In our first simulation experiment, we capture a $21\times21$ grid of images with $61\%$ overlap between neighboring images.
Each captured image is $512$ px $\times512$ px, but high frequency information has been lost due to the bandpass filter.
Under this setup the Synthetic Aperture Ratio (SAR), given as the ratio between the synthetic aperture diameter and lens aperture diameter, is $8.8$.
\figurename~\ref{fig:simResChart}(b) shows the center observed image of the resolution target, in which features smaller than $12$ pixels are lost, due to the low-pass filtering of the aperture.
This corresponds to the best resolution that can be achieved in coherent illumination without using ptychography to recover additional frequency information.
If we use ptychography and phase retrieval, the resolution increases significantly; features as small as $2$ pixels can be recovered as shown in \figurename~\ref{fig:simResChart}(d).
Plots comparing the contrast of the image intensity for the center observation (dashed blue line) and the recovered image (solid purple line) are provided in \figurename~\ref{fig:simResChart}(c).
The recovered Fourier magnitude and zoomed in detail views of groups 2, 5, and 6 are shown in \figurename~\ref{fig:simResChart}(e) and (f) respectively, showing faithful recovery of otherwise unresolved features.

Both the cost of assembling the camera array and the computational complexity of our reconstruction algorithm is $O(N^2)$.
Therefore, minimizing the number of captured images necessary to recover each high resolution image is an important design criteria.
There are two degrees of freedom to explore to help reduce the amount of ptychographic data acquired: changing the amount of overlap between neighboring images and changing the SAR.

\begin{figure}[t!]%
\centering
\includegraphics[width=\columnwidth]{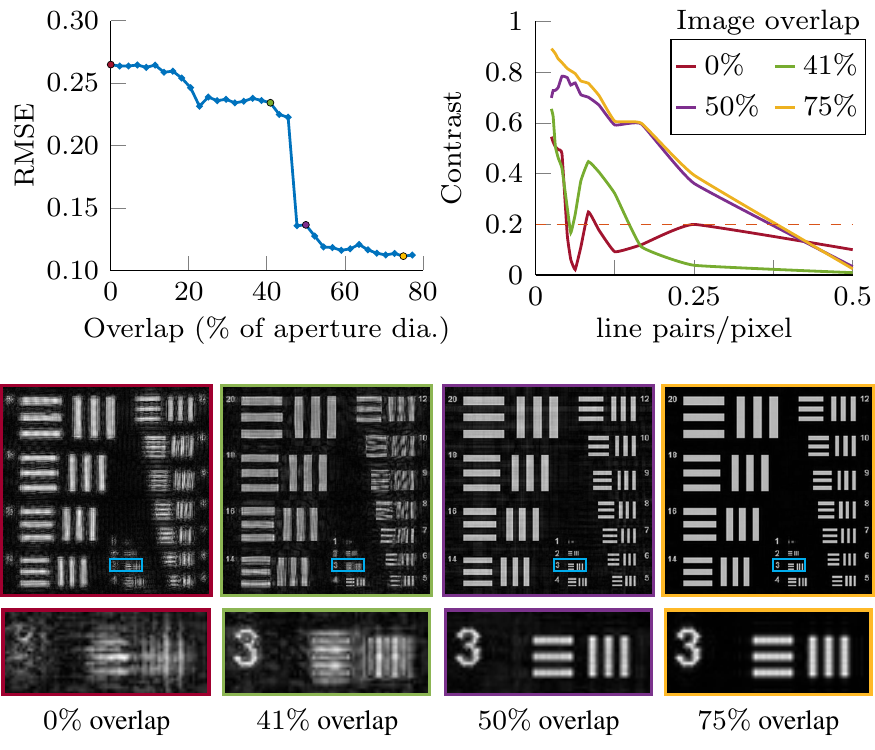}%
\caption{\textbf{Effect of varying overlap between adjacent images.} Holding the synthetic aperture size constant, we vary the amount of overlap between adjacent images. As the amount of overlap increases, reconstruction quality improves. When the amount of overlap is less than $50\%$, we are unable to faithfully recover the high resolution image, as shown in the RMSE plot in the top left. The contrast plots of the intensity images for four selected overlap amounts ($0\%,41\%,50\%,$ and $75\%$) shows that insufficient overlap drastically reduces the resolution of the system. Recovered images show the increase in image reconstruction quality as the amount of overlap increases.}%
\label{fig:varySpacing_resChart}%
\end{figure}

\Subsection{Effect of Overlap}
Intuitively, increasing the amount of overlap should improve reconstruction performance. 
Redundant measurements help to constrain the reconstruction and provide some robustness to noise.
In \figurename~\ref{fig:varySpacing_resChart}, we show the effects of increasing the amount of overlap from $0\%$ (no overlap) to $77\%$, a large degree of overlap. 
The SAR is held constant at $10$ for each overlap value, resulting in a $9\times9$ sampling grid for $0\%$ overlap and a much larger $41\times41$ grid for $77\%$ overlap.
As seen in \figurename~\ref{fig:varySpacing_resChart}, the root-mean-squared error (RMSE) between the ground truth and recovered intensities decreases as the amount of overlap increases with a precipitous drop when the amount of overlap is greater than $\sim\hspace{-1ex} 50\%$. 
This watershed demarcates the boundary of recovering the high resolution image.

We can see the effect of varying the amount of overlap in the contrast plot and the recovered images shown in \figurename~\ref{fig:varySpacing_resChart}.
Images captured with overlaps of $0\%$ and $41\%$ have large RMSE, fail to recover even low resolution features (even though the SAR is constant), and the resulting images are of low quality.
Conversely, data captured with overlaps of $50\%$ and $75\%$ are able to reconstruct small features with low RMSE.
Reconstruction quality also increases as the amount of overlap increases.
Thus $50\%$ is a lower bound on the amount of overlap required to run phase retrieval, with a preference for more overlap if application constraints permit.
This observation coincides with previously reported overlap factors of $60\%$ for good reconstruction~\cite{bunk2008influence}.

\begin{figure}[t!]%
\centering
\includegraphics[width=\columnwidth]{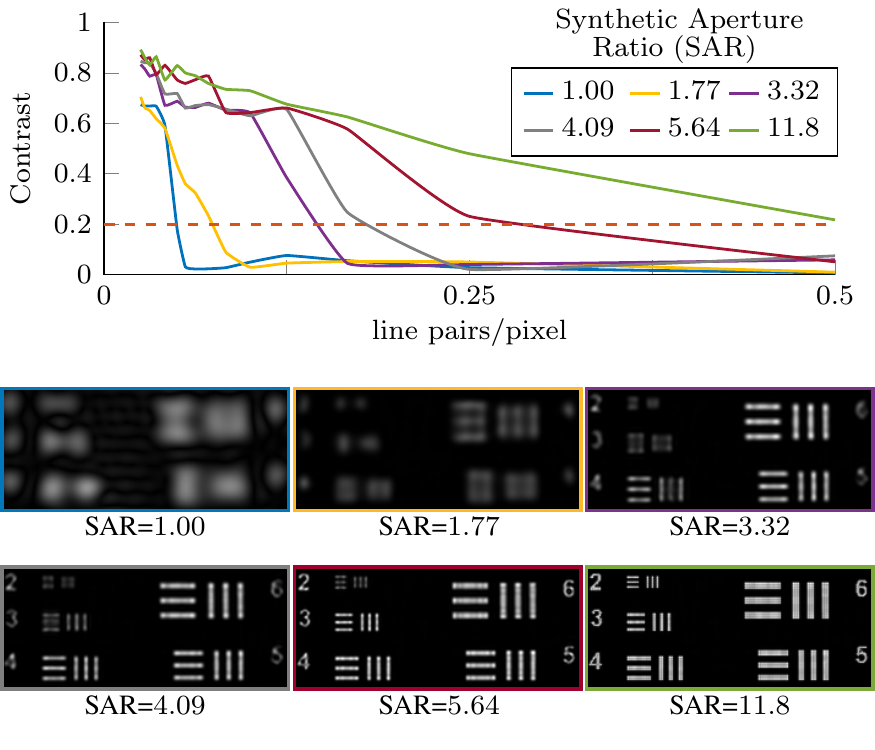}%
\caption{\textbf{Varying synthetic aperture ratio (SAR).} For a fixed overlap of $61\%$, we vary the size of the synthetic aperture by adding cameras to array. As seen in the contrast plot of image intensities, as well as the reconstructed intensities below, the resolution of the recovered images increase as the SAR increases from $1$ (the observed center image) to $11.8$. We are able to recover group $2$ with an SAR of $5.64$, which requires $169$ images. }%
\label{fig:varyN_resChart}%
\end{figure}

\Subsection{Effect of Synthetic Aperture Ratio (SAR)}
Alternatively, the number of images may be reduced by decreasing the synthetic aperture ratio.
For a given overlap, smaller SAR values will yield less improvement in resolution than larger SAR values (and a corresponding increase in the number of images).
To demonstrate this, we use the same camera parameters as the previous experiment and fix overlap at $61\%$ while varying the SAR.
\figurename~\ref{fig:varyN_resChart} shows that larger apertures can resolve smaller features, which can be clearly seen in the contrast plot.
Without using ptychography, features less than $12$ pixels wide ($0.04$ line pairs/pixel, $25~\mu$m at the image plane) cannot be resolved.
Resolution steadily improves as the SAR increases to $5.64$ before holding steady until an SAR of $11.8$.
As the SAR increases, image contrast also increases, which can be seen in the detail images shown in \figurename~\ref{fig:varyN_resChart}.
Of course, increasing the SAR requires additional images; an SAR of $1.77$ requires a $3\times3$ sampling grid, while SARs of $3.32$, $4.09$, $5.64$, and $11.8$ require $7\times7$, $9\times9$, $13\times13$, and $29\times29$ grids respectively.
While, in theory, coherent camera arrays can improve resolution down to the wavelength of the illumination source, an order of magnitude in SAR requires a quadratic increase in the number of recorded images.
Thus, the SAR should be set to the smallest size which meets the resolution requirement for a given application.

\begin{figure}[t!]%
\centering
\includegraphics[width=\columnwidth]{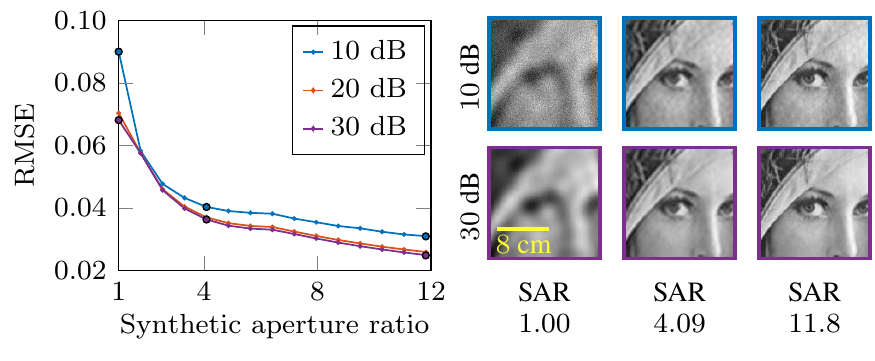}%
\caption{\textbf{Noise robustness of the coherent camera array.} We repeat the experiment in \figurename~\ref{fig:varyN_resChart} with the Lena image (assuming the scene is $650$ meters away), while varying the amount of added Gaussian noise. The signal-to-noise ratio (SNR) of the input images is set to be $10$, $20$, and $30$ dB. As shown in the RMSE plot, we are able to recover the high resolution image even with noisy input images. The observed center images (SAR=$1$) are blurry and noisy, with a high RMS error. As the SAR increases, the resolution improves, the noise is suppressed, and RMSE decreases. A cropped portion of select images are shown to highlight the performance of our approach.  Note: all other simulation experiments are conducted with an input SNR of $30$ dB.}%
\label{fig:varyN_lena}%
\end{figure}

\Subsection{Effect of Noise on Image Recovery}
The phase retrieval problem in \eqref{eq:problem1} is non-convex and thus prone to getting stuck in local minima when using gradient descent procedures such as alternating minimization.
Nonetheless, we observe that in practice our algorithm converges well for a variety of scenes and noise levels.
We repeat the varying SAR experiment using the canonical Lena image, and with varying levels of additive white Gaussian noise in the reconstruction.
We test signal-to-noise ratios (SNR) of $10$, $20$, and $30$ dB, and present the results in \figurename~\ref{fig:varyN_lena}.
As shown in the RMSE plot, we are able to recover the high resolution image even with input images containing significant noise.
The observed center images (SAR=$1$) reflect this noise, as well as diffraction blur, and thus exhibit a high RMS error.
As the SAR increases, the resolution improves, the noise is suppressed, and the RMSE decreases (thanks to redundancy in the measurements).
All remaining simulation experiments are conducted with an input SNR of $30$ dB.

\Subsection{Comparison with Expensive Large Aperture Lenses}
\label{sec:experimentalSim}
Next, we compare simulated images from an expensive and bulky lens, with simulated images from our coherent camera array. 
For these experiments we assume each system exhibits a fixed focal length of $1200$ mm.
In the early 1990s, Canon briefly made a $1200$ mm lens with a $215$ mm diameter aperture\footnote{Canon EF $1200$mm $f/5.6$L USM} which retailed for nearly \$$150{,}000$ US dollars (inflation adjusted to 2015) and weighed nearly $17$ kilograms. 
At $30$ meters, its minimum resolvable feature size is $190 \mu$m at the object plane (Nyquist-Shannon limit).
This diffraction spot size is just small enough to recover the pertinent features of a fingerprint where the average ridge width is on the order of $400~\mu$m \cite{orczyk2011fingerprint}.

Instead of using such a large and expensive lens, one could alternatively consider imaging with a much smaller lens with $1200$ mm focal length and a $75$ mm diameter which may cost on the order of hundreds of dollars to manufacture.
Such a lens would have a diffraction spot size of $21~\mu$m on the image sensor, and a minimum resolvable feature size of $536~\mu$m at $30$ meters.
The poor performance of this lens precludes identifying fingerprints directly from its raw images.
However, incorporating such a lens into a coherent camera array, capturing multiple images, and reconstructing with ptychographic phase retrieval can offer a final diffraction-limited resolution that surpasses the large Canon lens.

\begin{figure}[t!]%
\centering
\includegraphics[width=\linewidth]{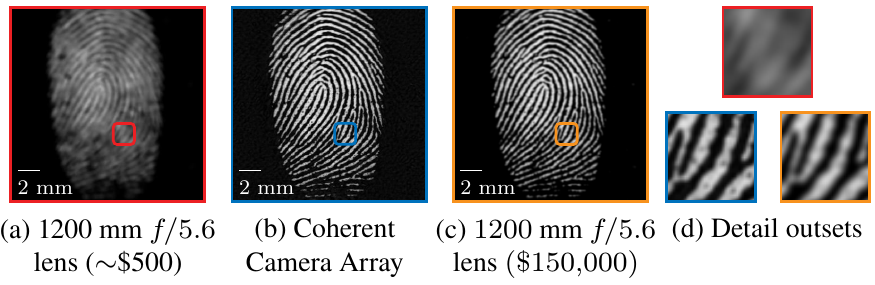}
\caption{\textbf{Simulation of imaging a fingerprint at $30$ meters.} We simulate image capture with an inexpensive $1200$ mm lens, $75$ mm aperture diameter ($f/16$), and a pixel pitch of $2~\mu$m. (a) Using passive illumination the resulting diffraction blur ($530~\mu$m) removes details necessary to identify the fingerprint. (b) Imaging with a coherent camera array ($61\%$ overlap) that captures $81$ images to create a synthetic aperture of $300$ mm reduces diffraction blur to $130~\mu$m, and leads to faithful recovery of minutiae. (c) Using a $150{,}000$ lens with a $215$ mm aperture diameter ($f/5.6$), the diffraction blur reduces to $190~\mu$m, which is roughly comparable to our final reconstruction in (b). (d) Detail views of the three imaging systems. 
In this simulation diffraction is the only source of blur, other factors in \tablename~\ref{tab:motivTab} are not considered.}%
\label{fig:simResults_fingerprint}%
\end{figure}

We simulate imaging a fingerprint at $30$ meters with both of the above approaches in \figurename~\ref{fig:simResults_fingerprint}
We use a fingerprint from the $2004$ Fingerprint Verification Challenge \cite{maltoni2009handbook} as ground truth, where each pixel corresponds to $51~\mu$m at the object plane, and the pixel size on the camera sensor is $2~\mu$m (as before).
Using a $1200$ mm focal length, $75$ mm aperture lens with incoherent illumination yields images in which blur has degraded the image quality so that they are of no practical use (\figurename~\ref{fig:simResults_fingerprint}(a)).
However, if the same lens is used to simulate an aperture diameter of $300$ mm in a coherent camera array ($61\%$ overlap, $9\times9$ images, SAR=$4$), the diffraction spot size reduces to $130~\mu$m, which is sufficient to identify the print, as shown in \figurename~\ref{fig:simResults_fingerprint}(b).
The Canon lens with a $215$ mm aperture could be used to achieve comparable quality, as in \figurename~\ref{fig:simResults_fingerprint}(c).
Detail images in \figurename~\ref{fig:simResults_fingerprint}(d) highlight the image quality of the three imaging systems.

\begin{figure}[t!]%
\centering
\includegraphics[width=\linewidth]{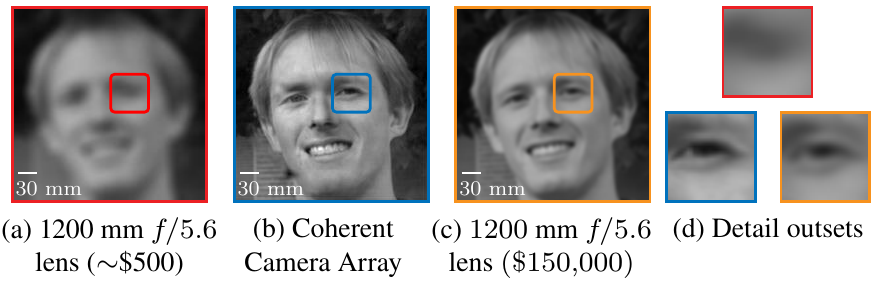}
\caption{\textbf{Simulation of capturing a face $1000$ meters away.} We simulate image capture with an inexpensive $1200$ mm lens, $75$ mm aperture diameter ($f/16$), and a pixel pitch of $2~\mu$m. (a) Directly using the inexpensive $f/16$ lens in passive illumination results in a diffraction blur spot size of $17.8$ mm on the face, obliterating detail necessary to recognize the subject. (b) Using Fourier ptychography ($61\%$ overlap, $81$ images) to achieve a synthetic aperture of $300$ mm the diffraction spot size is reduced to $4.4$ mm. (c) Using the $\$150{,}000$ $215$ mm ($f/5.6$) lens yields a diffraction spot size of $6.2$ mm, $50\%$ larger than the diffraction realized using Fourier ptychography. Detail views of the three systems are shown in (d). In this simulation diffraction is the only source of blur, other factors in \tablename~\ref{tab:motivTab} are not considered.}%
\label{fig:simResults_face}%
\end{figure}

A similar experiment is conducted in \figurename~\ref{fig:simResults_face} with the subject being a human face $1000$ meters away.
In this experiment, a synthetic aperture of $300$ mm, corresponds to a diffraction spot size of $4.4$ mm on the face.
The $75$ mm lens has a diffraction spot size of $180$ mm and is again unable to resolve the scene with incoherent illumination but fine details are recovered using ptychography, \figurename~\ref{fig:simResults_face}(a) and (b) respectively.
The Canon lens (\figurename~\ref{fig:simResults_fingerprint}(c)), yields a blurry image that is inferior to the image acquired under active illumination.
Detailed crops of one of the eyes acquired using the three imaging setups are shown in \figurename~\ref{fig:simResults_face}(d).

While these simulations are limited in scope (i.e. we assume that diffraction is the only factor limiting resolution and neglect phase), the results suggest that implementing Fourier ptychography in a camera array offers a practical means to acquire high resolution images while remaining cost competitive with existing lenses.
Moreover, such computational approaches appear as the only means available to surpass the resolution limits imposed by current manufacturing technology.

%%%%%
\Section{Experimental Results}
In this section, we experimentally verify our simulations of the coherent camera array. 
For simplicity, we use a transmissive setup similar to existing ptychography methods applied in microscopy.
While this configuration is not practical for long distance imaging, we present the results as a proof-of-concept that we hope will directly extend to a reflection geometry.
In Section~\ref{sec:prelimWork} we discuss the practical limitations of implementing a long-range ptychography imaging system

\begin{figure}[t!]%
\centering
\includegraphics[width=3.3in]{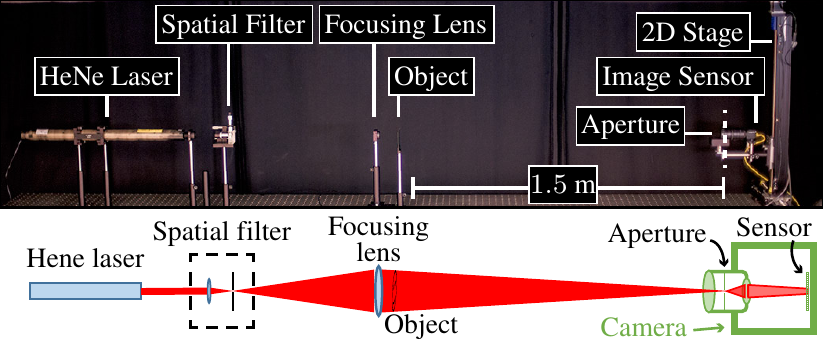}%
\caption{\textbf{Overview of hardware configuration for data acquisition (side view).} From left to right: A helium neon laser passed through a spatial filter acts as the coherent illumination source. A focusing lens forms the Fourier transform of the transmissive object at the aperture plane of the camera's lens. The aperture acts as a bandpass filter of the Fourier transform and the signal undergoes an inverse Fourier transform as it is focused onto the camera's sensor. The camera (Point Grey Blackfly (BFLY-PGE-50A2M-CS) is mounted on a motorized $2$D stage to capture overlapping images.}%
\label{fig:setup}%
\end{figure}

\Subsection{Experimental Prototype}
Our imaging system consists of a Blackfly machine vision camera manufactured by Point Grey (BFLY-PGE-50A2M-CS), coupled with a $75$ mm Fujinon lens (HF75SA-1) that is mounted on a motorized $2$D stage (VMX Bi-slide, see \figurename~\ref{fig:setup}).
For a single acquired data set, we scan over the synthetic aperture in a raster pattern. We discuss future geometries that avoid physical scanning in the next section.
Objects to be imaged are placed $1.5$ meters away from the camera and are illuminated with a helium neon (HeNe) laser which emits a beam with a wavelength of $633$ nm.
We employ a spatial filter to make the laser intensity more uniform as well as to ensure sufficient illumination coverage on our objects.
A lens is placed immediately before the target to collect the illumination and form the Fourier transform on the aperture plane of the camera lens.

Lenses with small focal lengths can be easily (and cheaply) manufactured to have a proportionally large aperture. The maximum aperture of the lens that we use is $42$ mm.
In order to simulate building a cheap lens with a long focal length and small aperture, we stopped the lens down to $f/32$, creating a $2.3$ mm diameter aperture.
The diffraction spot size on the sensor is $49~\mu$m in diameter.
Given that the camera sensor has a pixel width of $2.2~\mu$m we expect to see a blur of roughly $20$ pixels.

Unique images are acquired by moving the camera to equidistant positions in the synthetic aperture plane via the translation stage.
At every grid position, we capture multiple images with different exposures to form a high dynamic range reconstruction.
By translating the camera, the shift in perspective causes the captured images to be misaligned.
For planar scenes a pair of misaligned images can be corrected by finding the homography relating the reference viewpoint (taken to be the center position) and the outlying view.
We accomplish this by finding fiducial markers of a checkerboard pattern affixed to the object, which is lit with incoherent light.
The translation stage is of high enough precision to allow the calibration step to be performed immediately before or after data capture.
The aligned high dynamic range images are used as inputs to the phase retrieval algorithm.

\begin{figure*}[t]%
\centering
\includegraphics[width=\linewidth]{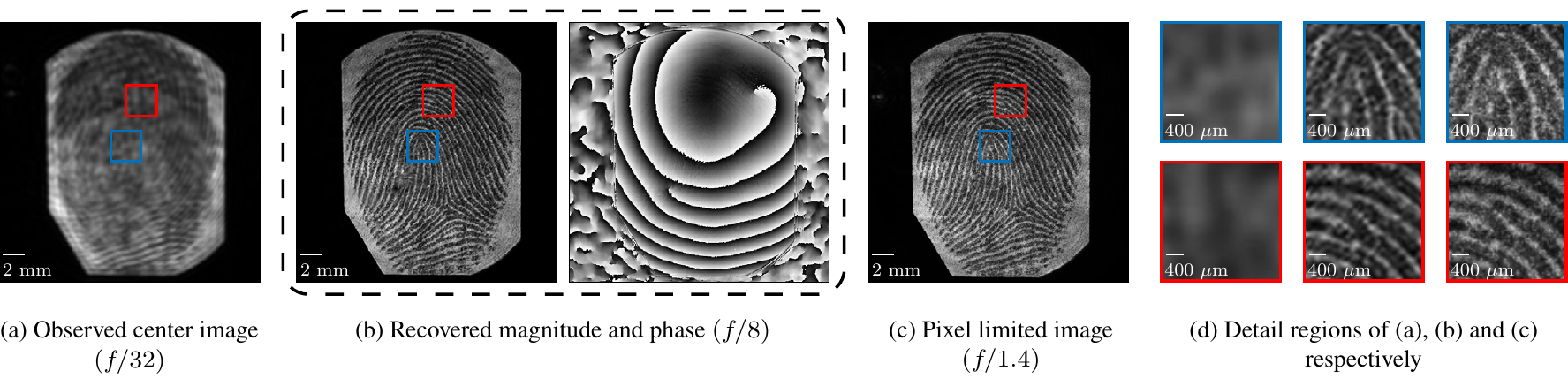}
\caption{\textbf{Experimental result: Resolving a fingerprint $4\times$ beyond the diffraction limit.} Using the hardware setup shown in \figurename~\ref{fig:setup}, we use a $75$ mm focal length lens to acquire images of a fingerprint $\sim 1.5$ m away from the camera. A fingerprint was pressed on a glass slide and fingerprint powder was used to make the ridges opaque. (a) Stopping down the aperture to a diameter of $2.34$ mm induces a diffraction spot size of $49~\mu$m on the sensor ($\sim 20$ pixels) and $990~\mu$m on the object. (b) We record a grid of $17\times17$ images with an overlap of $81\%$, resulting in a SAR which is $4$ times larger than the lens aperture. After running phase retrieval, we recover a high resolution magnitude image and the phase of the objects. (c) For comparison, we open the aperture to a diameter of $41$ mm to reduce the diffraction blur to the size of a pixel. (d) Comparing zoomed in regions of the observed, recovered, and comparison images shows that our framework is able to recover details which are completely lost in a diffraction limited system. }%
\label{fig:realFingerprint}%
\end{figure*}

\begin{figure*}[t]%
\centering
\includegraphics[width=\linewidth]{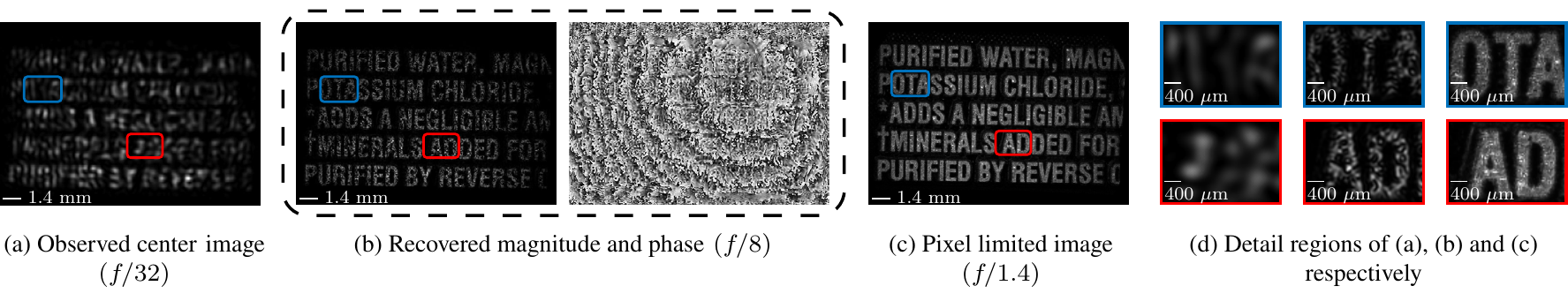}
\caption{\textbf{Experimental result: Resolving $4\times$ beyond the diffraction limit for a diffuse water bottle label.} Using the same parameters as in \figurename~\ref{fig:realFingerprint}, we image a diffuse water bottle label $\sim 1.5$ m away from the camera. The diffuse nature of the water bottle label results in laser speckle. In the observed center image (a), diffraction blur and laser speckle render the text illegible. Using Fourier ptychography (b) we are able to reduce the effect of speckle and remove diffraction revealing the text. In the comparison image (c) the text is clearly legible. Detail views in (d) show the improvement in image resolution when using Fourier ptychography.}%
\label{fig:realDasani}%
\end{figure*}

\begin{figure*}[t]%
\centering
\includegraphics[width=\linewidth]{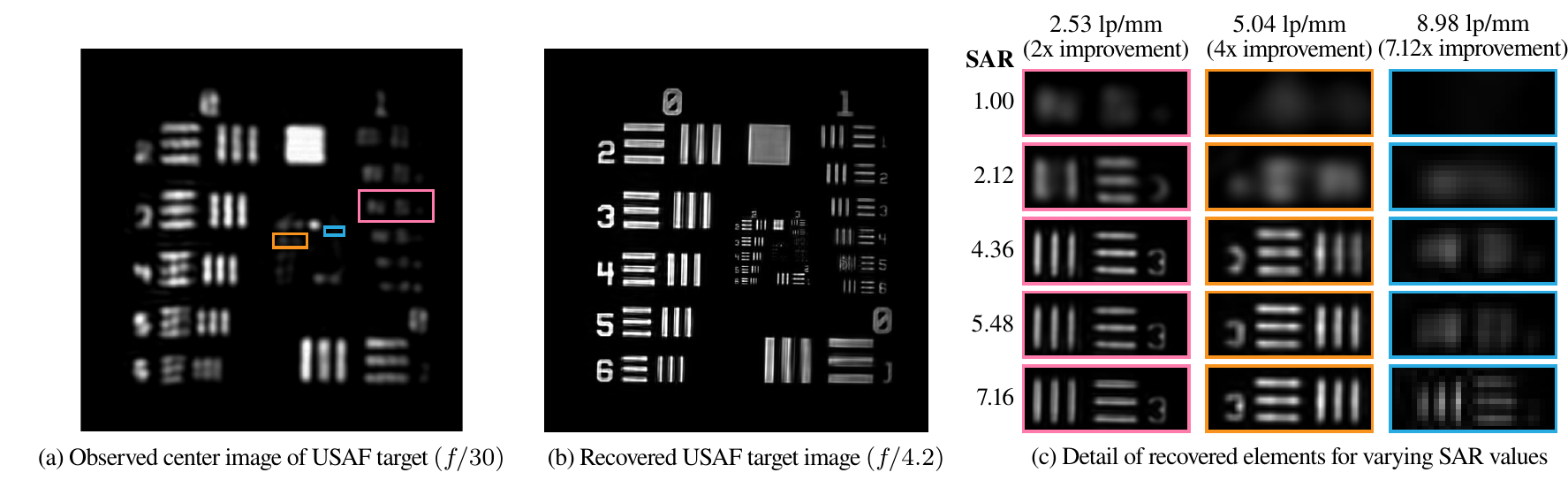}
\caption{\textbf{Experimental result: Recovering a USAF target with varying SAR.} We capture a USAF resolution target $1.5$ meters away from the camera. For this experiment, we use a slightly different lens and image sensor (focal length = $75$ mm, pixel pitch = $3.75~\mu$m). (a) The camera is stopped down to $2.5$ mm which induces a $930~\mu$m blur on the resolution chart, limiting resolution to $1.26$ line pairs per millimeter (lp/mm). (b) We record a $23\times23$ grid of images with $72\%$ overlap, resulting in a SAR of $7.16$. Following phase retrieval, we are able to resolve features as small as $8.98$ lp/mm, a $7.12\times$ improvement in resolution. (c) We show the effect of varying the SAR on resolution. Using a subset of the captured images we vary the SAR from $2.12$ up to $7.16$. The gains in improvement closely track the SAR.}%
\label{fig:USAF}%
\end{figure*}

\Subsection{Recovered images}
We test our setup with three scenes. First, we deposit a fingerprint on a glass microscope slide and use consumer grade fingerprint powder to reveal the latent print. Our second scene contains a translucent water bottle label, and our third scene is a resolution target.
For the scenes containing the fingerprint and water bottle labels, we record a $17 \times 17$ grid of images with $81\%$ overlap between adjacent images, resulting in a synthetic aperture with a $9.3$ mm diameter (SAR=$4$).

\textbf{Fingerprint:} \figurename~\ref{fig:realFingerprint} shows the results of using our coherent camera array prototype to improve the resolution of a fingerprint.
With an aperture diameter of $2.3$ mm, little high spatial frequency information is transferred to the image plane in a single image, resulting in the blurry image in \figurename~\ref{fig:realFingerprint}(a).
The coherent camera array technique recovers a high resolution Fourier representation of the object, from which we recover both the high resolution image and the phase of the object (\figurename~\ref{fig:realFingerprint}(b)).
The wrapping in the recovered phase could be caused by the illumination wavefront, or by a tilt in the object with respect to the camera.
One consequence of stopping down the lens aperture is that we are able to fully open the aperture and record and image which does not exhibit any diffraction blur, \figurename~\ref{fig:realFingerprint}(c), which provides a useful comparison for our reconstruction.
Detail views of the three imaging scenarios are shown in \figurename~\ref{fig:realFingerprint}(d). 

\textbf{Dasani label:} We next use the same setup to capture a translucent Dasani water bottle label, shown in \figurename~\ref{fig:realDasani}.
Unlike the fingerprint, the diffuse water bottle label has a random phase which produces a highly varying optical field at the image plane, which is reminiscent of laser speckle.
Speckle becomes more pronounced for small apertures (\figurename~\ref{fig:realDasani}(a)).
The recovered image in \figurename~\ref{fig:realDasani}(b) shows that the coherent camera array can recover details that are completely lost in a diffraction limited system, and the phase retrieval algorithm is able to handle the quickly varying random phase of the label.
Even with a modest increase to the SAR, individual letters are legible, as highlighted in the detailed crops in \figurename~\ref{fig:realDasani}(d). 
If we were to further increase the SAR, we could further reduce speckle-induced artifacts, as suggested by the raw open-aperture image in \figurename~\ref{fig:realDasani}(c).

\textbf{USAF target:} We use the same setup to capture a USAF target, but now with a different $75$ mm lens (Edmund Optics \#54-691) and a slightly different Blackfly camera (BFLY-PGE-13S2M-CS) with a pixel pitch of $3.75~\mu$m.
Using this setup, and an aperture diameter of $2.5$ mm, we expect a diffraction blur size of $48~\mu$m on the image sensor (approximately $12$ sensor pixels).
With the USAF target placed $1.5$ meters away from the camera, we expect a diffraction blur size of $970~\mu$m at the object plane.
We observe that the resolution of the USAF target is limited to $1.26$ line pairs per millimeter (lp/mm) in the center observed image (Group 0, Element 3 in \figurename~\ref{fig:USAF}a).
By acquiring a $23\times23$ grid of images, with $72\%$ overlap between neighboring images, we achieve an effective synthetic aperture ratio of $7.16$.
We show the results of such a coherent camera array reconstruction in \figurename~\ref{fig:USAF}(b), where the spatial recovered spatial resolution has increased to $8.98$ lp/mm (Group 3, Element 2), a $7.12\times$ improvement in resolution.

As discussed in Section~\ref{sec:simRecov}, varying the synthetic aperture ratio impacts the system resolution enhancement factor. 
This can be observed by varying the number of images used to recover the USAF target.
In \figurename~\ref{fig:USAF}(c) we present two regions of the target for a varying SAR of $1-7.16$.
As the SAR increases to $2.12$, $4.36$, $5.48$, and $7.16$ the realized resolution gains are $1.94$, $4.00$, $5.66$, and $7.12$ respectively, which closely tracks the expected behavior.

\textbf{Reproducible Research:} We are committed to reproducible research. Our codes and data sets may be found on our project webpage~\cite{projectWebpage} (\href{http://jrholloway.com/projects/towardCCA}{http://jrholloway.com/projects/towardCCA}).

%%%%%
\Section{Barriers to Building a Real System}
\label{sec:prelimWork}
In this section we discuss potential barriers to implementing a coherent camera array for long range imaging.
Where possible, we present preliminary simulation results.

\textbf{Single-shot imaging:} One limitation of the ptychography framework with a single camera and illumination source is that it assumes a static or a slow-moving scene. 
This restriction arises because, in order to sample a large Fourier space, we capture images while moving the camera one step at a time. 

\begin{figure}[t!]%
	\centering
	\includegraphics[width=\columnwidth]{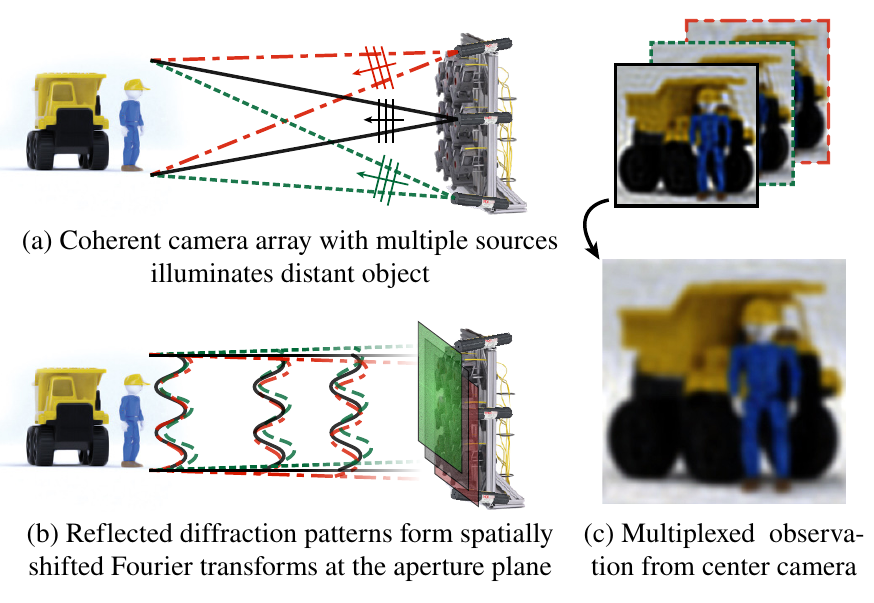}
	\caption{\textbf{An illustration of ptychography with multiplexed illumination and a camera array.}%
	(a) Multiple coherent sources illuminate a distant scene. (Note: different colors are used to distinguish light sources; however, we assume each source operates in the same wavelength.) (b) Reflected light undergoes Fraunhofer diffraction forming spatially shifted Fourier transforms at the aperture plane of the camera array. (c) A single camera aperture (e.g. center camera) sums the intensities from each diffraction pattern (top), recording a single image (bottom).}
	\label{fig:mux-illustration}%
\end{figure}

To enable imaging dynamic scenes, we envision a design with multiple cameras and illumination sources that can sample a large Fourier space in a \textit{single shot}. We illustrate the modified design in \figurename~\ref{fig:mux-illustration}, where we consider an array of cameras in a tightly packed configuration and an array of illumination sources that are placed such that the optical fields generated by them at the aperture plane are slightly shifted with respect to one another. 

By changing the location and angle of the illumination source, we can shift the optical field at the aperture plane. Therefore, if we use multiple illumination sources at once, they will produce multiple, shifted optical fields at the aperture plane. 
In such a multiplexed illumination setting, each camera in the array samples multiple, shifted copies of a reference Fourier plane. The sampling pattern is determined by the relative positions and orientations of the cameras and the illumination sources. 
Since the illumination sources are incoherent with each other, the intensity image measured at each camera corresponds to the sum of intensities from different Fourier regions. 
Furthermore, we can also capture multiple images by changing the illumination pattern. Similar multiplexed illumination methods have already been used for high-resolution microscopy in~\cite{tian2014multiplexed}. 

Suppose our system consists of $P$ cameras and $Q$ illumination sources that divide the Fourier space into overlapping apertures with centers $(c_{x'_i},c_{y'_i})$, where $i$ indicates a camera-illumination source pair.
Suppose we capture $T$ images while changing the illumination patterns. Let us describe the images measured by camera $p\in\{1,\ldots,P\}$ with the $t$th illumination pattern as follows
\begin{equation}\label{eq:mux_Model}
I_{p,t} = \sum_{i\in \mathcal{M}(p,t)}|\mathcal{F}\mathcal{R}_{\Omega_i}\widehat \psi|^2,
\end{equation}
where $\mathcal{M}(p,t)$ denotes the indices for those apertures in the Fourier plane that are multiplexed in the recorded image $I_{p,t}$. 
$\widehat \psi$ denotes the complex-valued field at the aperture plane and $\mathcal{R}_{\Omega_i} \widehat \psi$ denotes an aperture operator that only selects the entries of $\widehat \psi$ inside the set $\Omega_i$. 
To recover the high-resolution $\widehat\psi$ from a sequence of $P*T$ multiplexed, low-resolution, intensity measurements, $\{I_{p,t}\}$, we use an alternating minimization-based phase retrieval problem similar to the one depicted in Figure~\ref{fig:AM_blocks}, where we iteratively estimate the complex-valued field at the sensor planes: $\psi_i^k = \mathcal{F}\mathcal{R}_{\Omega_i} \widehat \psi$.
The only change appears in enforcing the magnitude constraints, where we replace the magnitudes of $\psi_{i}^k$ at $k$th iteration as follows,
\begin{equation}\label{eq:mux_magnitudeReplacement}
 \psi_{i}^k \gets \sqrt{\frac{I_{k,t}}{\sum_{i\in\mathcal{M}(k,t)}|\psi_i^k|^2}}\; \psi_i^k \quad \text{for all } i \in \mathcal{M}(p,t).
\end{equation}

\begin{figure}[t!]
	\centering
	\includegraphics[width=\linewidth]{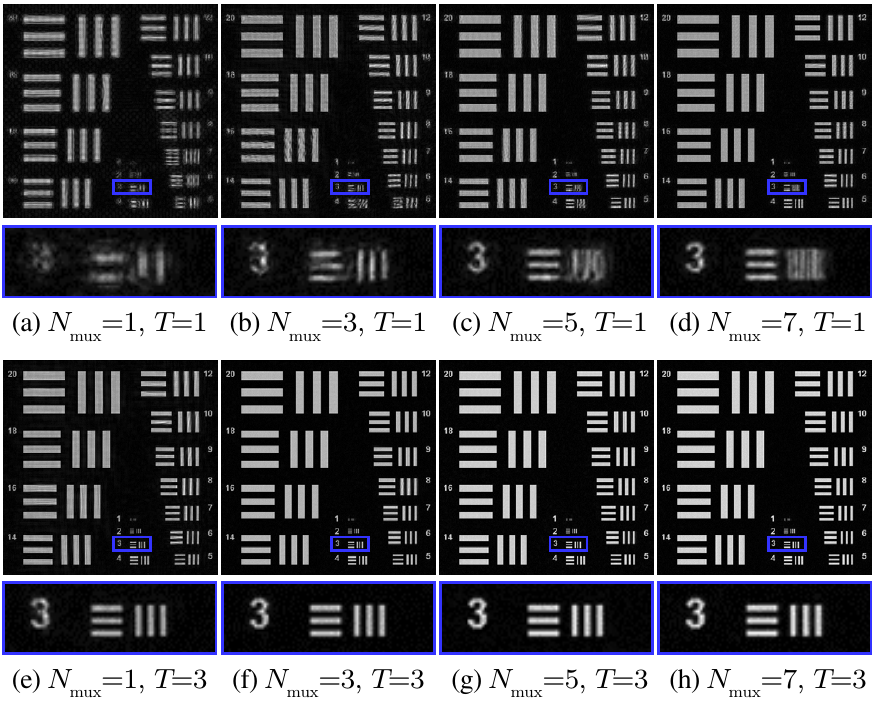}
	\caption{\textbf{Simulation of recovering a resolution target using multiplexed illumination patterns.} We repeat the conditions of \figurename~\ref{fig:simResChart} using a $7\times7$ camera array. Each camera has an $800$ mm lens with an aperture (front lens diameter) of $25$ mm. Cameras are placed such that their apertures abut, though don't overlap, in the Fourier plane. Illumination sources provide $66\%$ overlap among neighboring apertures in the Fourier plane. $N_\text{mux},T$  in subcaptions denote the number of active illumination sources per image ($N_\text{mux}$) and the number of images captured with different illumination patterns ($T$), respectively. Recovered images show that increasing $N_\text{mux}$ and $T$ improves the quality of the reconstructed images. }\label{fig:mux-results}
\end{figure}

To gauge the potential of the multiplexed illumination scheme, we simulate the recovery of a resolution chart using a $7\times 7$ camera array.
Cameras are positioned in a square grid such that their lenses abut; each camera effectively samples a non-overlapping region of the Fourier plane.
Every camera is assumed to have a lens with an $800$ mm focal length, and an aperture limited by the front lens diameter of $25$ mm.
The resolution chart ($64$ mm $\times 64$ mm) is placed $50$ meters away, and assuming a light source with a wavelength of $550$ nm and a pixel pitch of $2~\mu$m, the diffraction spot size is $\approx 20$ pixels.
We use a $7\times 7$ array of illumination sources adjusted so that they provide $66\%$ overlap between adjacent apertures in the Fourier plane. 

We present the results in \figurename~\ref{fig:mux-results}, where we examine the effects of increasing the number of active illumination sources per image ($N_\text{mux}$) and the number of images captured with different illumination patterns ($T$). 
We recorded multiplexed images under different illumination patterns according to the model in \eqref{eq:mux_Model}. 
In each experiment, we recorded $T$ images by randomly selecting a different $N_\text{mux}$ active illumination sources from the $7\times 7$ for every image. 
We recovered the resolution target using the algorithm described in \figurename~\ref{fig:AM_blocks} along with the modification in \eqref{eq:mux_magnitudeReplacement}.
The sampling scheme with $N_\text{mux}=1$, $T=1$ (i.e. ptychography with $0\%$ overlap) leaves holes in the Fourier plane; therefore, the reconstruction shows severe aliasing artifacts.
As we increase $N_\text{mux}$, the sampling space fills up the entire Fourier plane and also introduces overlaps in the observed measurements, which improves the reconstruction quality.
\figurename~\ref{fig:mux-results}(a)--\ref{fig:mux-results}(d) show the results for single-shot imaging ($T=1$) with increasing number of active illumination sources per image. 
The quality of reconstruction significantly improves if we capture multiple images with the camera array. 
\figurename~\ref{fig:mux-results}(e)--\ref{fig:mux-results}(h) show the images recovered from $T=3$ sets of images, each captured with a different pattern of $N_\text{mux}$ active illumination sources. 
These results indicate that combining a camera array with an array of programmable illumination sources can help us design a single- or multi-shot imaging system that can potentially capture real-world, dynamic scenes. 

\textbf{Diffuse Materials:} Diffuse objects are materials which have a rough surface at the scale of illumination wavelength.
The randomness in the surface manifests as a rapidly changing random phase for the object.
In \figurename~\ref{fig:realDasani}, we presented results for a diffuse water bottle label which exhibited significant laser speckle due to the random phase of the label.
We were able to recover the high-resolution magnitude and phase.
While this material was weakly diffuse, it suggests that it should be possible to use Fourier ptychography to improve resolution of everyday objects in long distance images.

\textbf{Reflective Recovery:} Traditionally, ptychography has been restricted to working in a transmissive modality.
This arrangement is impractical in long range imaging, if one can get close enough to position a laser to shine through an object then they can snap a picture.
Recent work suggests that Fourier ptychography can be extended to work in reflective mode~\cite{liang2015transfer}, though more research is needed to characterize the performance and confirm it as a viable technique for long distance imaging.

%%%%%
\Section{Conclusion}
\label{sec:conclusion}
In this paper we have proposed using Fourier ptychography to dramatically improve resolution in long distance imaging.
Whereas existing super-resolution techniques only offer a $2\times$ improvement in resolution, the gains in ptychography are (theoretically) orders of magnitude greater.
We have demonstrated resolution gains of $7.12\times$ for real scenes and show simulation results with $10\times$ improvement in resolution.

One consideration which must be addressed before realizing a long-distance ptychographic system is how to account for dynamic scenes. 
In the current system, a camera is affixed to a translation stage and images are acquired sequentially over the course of tens of minutes.
We have proposed using an array of cameras fitted with inexpensive lenses and multiplexed illumination to capture images simultaneously, a necessary requirement for capturing dynamic scenes.
A greater concern is implementing a ptychographic system in reflective mode.
It is impractical to position a laser source behind an object to be imaged remotely, which necessitates the placement of the illumination source near to the image sensor.
We believe our initial results show great promise that a practical reflective mode ptychography system can be implemented for imaging long distances.
In the future, we plan to design, build, and test such camera array systems as an extension of the work presented here.

{\small
\bibliographystyle{ieee}
\bibliography{towardCCA}
}

% that's all folks
\end{document}